\documentclass{article} 
\usepackage{iclr2027_conference,times}

\usepackage{amsmath,amsfonts,bm}

\def\eqref#1{equation~\ref{#1}}

\def\1{\bm{1}}

\DeclareMathAlphabet{\mathsfit}{\encodingdefault}{\sfdefault}{m}{sl}
\SetMathAlphabet{\mathsfit}{bold}{\encodingdefault}{\sfdefault}{bx}{n}

\usepackage{hyperref}
\usepackage{url}
\usepackage{graphicx}
\usepackage{booktabs}
\usepackage{tikz}
\usetikzlibrary{arrows.meta,calc,positioning}

\usepackage{colortbl}

\usepackage{hyperref}
\usepackage{url}
\usepackage{booktabs}
\usepackage{multirow}
\usepackage{xcolor}
\usepackage{tikz}
\usepackage{enumitem}
\usepackage{algorithm}
\usepackage{algpseudocode}
\definecolor{brightcerulean}{rgb}{0.11, 0.67, 0.84}

\usepackage{threeparttable}
\usepackage{longtable}
\usepackage{array} 
\usepackage{tcolorbox}
\newcommand{\methodname}{\texttt{MTA-Zero}}

\title{\methodname: Self-Improving Multimodal Deep Search via Online
QA--Solver Co-training}

\author{Xiangyu Peng, Jiaxin Zhang, An Yan, Yu Li \\
Ran Xu, Zeyuan Chen, Chien-Sheng Wu \\
\texttt{becky.peng@salesforce.com}}
\iclrfinalcopy
\begin{document}

\maketitle

\begin{abstract}
Multimodal large language models (MLLMs) have become strong visual reasoners, yet remain limited in open-world, multi-step search that requires grounding images, selecting tools, and integrating web evidence. 
Existing approaches typically construct and filter fixed offline corpora whose question distributions cannot adapt as the target deep-search agent improves. 
Building on MTA-Agent's image-grounded QA construction~\citep{peng2026mta}, we present \methodname{}, an online multimodal QA--Solver co-training framework for self-improving multimodal deep-search agents that requires no precomputed multi-hop corpus. Starting from image-grounded VQA seed contexts, a trainable QA Generator retrieves evidence and constructs verified multi-hop questions online. A separate deep-search Solver then selects tools, gathers evidence, and answers the generated questions. To focus generation on the Solver's learning frontier, we derive a Goldilocks reward for the QA Generator from the Solver's group-wise rollout success rate: questions solved by all rollouts or by none receive zero reward, whereas questions near the current frontier are favored. We optimize the two policies independently with GRPO, without sharing parameters or gradients. Across six multimodal deep-search benchmarks, \methodname{} improves a Qwen3-VL-8B Solver from $36.91\%$ to $51.58\%$ mean accuracy and achieves performance comparable to GPT-5 under the same agent workflow.  The same framework also improves a newer Qwen3.6-27B Solver by $10.82$ points. These results show that verified online question generation provides a policy-specific curriculum for continually improving multimodal deep-search agents.
\end{abstract}

\section{Introduction}
\label{sec:introduction}

Multimodal large language models (MLLMs) have made substantial progress in
visual understanding and reasoning~\citep{xue2024xgen,deepseekai2025deepseekv32,huang2025vision,qwen3technicalreport}. Yet, turning them into open-world agents
that can identify entities in images, gather web evidence, and compose
multi-hop answers through external tools remains challenging~\citep{geng2025webwatcher,chng2025sensenova,wu2025mmsearch}. 
Improving a target multimodal deep-search agent remains an open problem. We
refer to this deployment-time agent as the \emph{Solver agent}. 
The Solver must learn not only to reason over an image, but also
to decide what to search for, which tools to invoke, and how to integrate
evidence collected across modalities.

A dominant approach is to improve multimodal deep-search agents through an offline data
factory \citep{peng2026mta,zhang2026vsearcher,yao2026mm,chen2026opensearch}.
Starting from seed images or questions, capable models and tools
synthesize candidate multi-hop QA pairs and search trajectories; verification
and filtering then turn them into SFT or RL corpora such as MTA-Agent
\citep{peng2026mta}, MM-DeepResearch \citep{yao2026mm}, and REDSearcher
\citep{chu2026redsearcher}. 
These pipelines create
valuable, broadly reusable datasets, but their question distributions are fixed before the target Solver begins to evolve. 
As the Solver evolves,
a generic question may be unanswerable at an early stage, trivial at a later
one, or simply unrelated to its current capability gaps. 
Consequently, a fixed offline corpus may devote much of its generation,
verification, and tool-interaction workload to questions that are no longer
informative for the evolving policy.

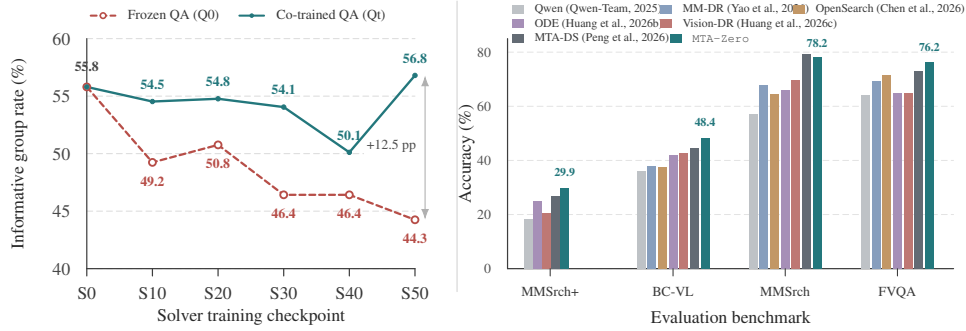
\begin{figure*}[t]
  \centering
  \resizebox{0.92\textwidth}{!}{
\begin{tikzpicture}[x=1cm,y=1cm,font=\rmfamily,line cap=round,line join=round]
  \definecolor{frozenqa}{RGB}{183,78,72}
  \definecolor{onlineqa}{RGB}{35,119,124}
  \definecolor{qwenbar}{RGB}{190,194,198}
  \definecolor{mmdrbar}{RGB}{135,158,189}
  \definecolor{openbar}{RGB}{194,151,102}
  \definecolor{odebar}{RGB}{166,143,183}
  \definecolor{visionbar}{RGB}{190,120,116}
  \definecolor{mtabar}{RGB}{99,107,116}

  \foreach \y/\ticklabel in {1.25/40,2.275/45,3.30/50,4.325/55,5.35/60} {
    \draw[gray!28,densely dashed,line width=.32pt] (1.75,\y) -- (7.60,\y);
    \node[anchor=east,font=\rmfamily\small,text=black!72] at (1.58,\y) {\ticklabel};
  }
  \draw[black!76,line width=.62pt] (1.75,1.25) -- (7.60,1.25);
  \draw[black!76,line width=.62pt] (1.75,1.25) -- (1.75,5.35);
  \foreach \x/\ticklabel in {1.75/S0,2.92/S10,4.09/S20,5.26/S30,6.43/S40,7.60/S50} {
    \draw[black!76,line width=.62pt] (\x,1.25) -- (\x,1.12);
    \node[anchor=north,font=\rmfamily\small,text=black!78] at (\x,1.05) {\ticklabel};
  }
  \node[font=\rmfamily\small,text=black!84] at (4.68,.39) {Solver training checkpoint};
  \node[rotate=90,font=\rmfamily\small,text=black!84] at (.55,3.30)
    {Informative group rate (\%)};

  \coordinate (szero) at (1.75,4.489);
  \coordinate (frozenTen) at (2.92,3.144);
  \coordinate (frozenTwenty) at (4.09,3.456);
  \coordinate (frozenThirty) at (5.26,2.566);
  \coordinate (frozenForty) at (6.43,2.566);
  \coordinate (frozenFifty) at (7.60,2.121);
  \coordinate (onlineTen) at (2.92,4.229);
  \coordinate (onlineTwenty) at (4.09,4.278);
  \coordinate (onlineThirty) at (5.26,4.130);
  \coordinate (onlineForty) at (6.43,3.323);
  \coordinate (onlineFifty) at (7.60,4.694);
  \draw[frozenqa,densely dashed,line width=1.15pt]
    (szero) -- (frozenTen) -- (frozenTwenty) -- (frozenThirty) --
    (frozenForty) -- (frozenFifty);
  \draw[onlineqa,line width=1.20pt]
    (szero) -- (onlineTen) -- (onlineTwenty) -- (onlineThirty) --
    (onlineForty) -- (onlineFifty);
  \foreach \point in {szero,frozenTen,frozenTwenty,frozenThirty,frozenForty,frozenFifty} {
    \draw[frozenqa,fill=white,line width=.95pt] (\point) circle[radius=1.75pt];
  }
  \foreach \point in {szero,onlineTen,onlineTwenty,onlineThirty,onlineForty,onlineFifty} {
    \filldraw[onlineqa,draw=onlineqa,line width=.45pt] (\point) circle[radius=1.12pt];
  }

  \draw[frozenqa,densely dashed,line width=1.05pt] (1.82,5.76) -- (2.26,5.76);
  \draw[frozenqa,fill=white,line width=.82pt] (2.04,5.76) circle[radius=1.40pt];
  \node[anchor=west,font=\rmfamily\scriptsize,text=black!78] at (2.36,5.76)
    {Frozen QA (Q0)};
  \draw[onlineqa,line width=1.05pt] (4.47,5.76) -- (4.91,5.76);
  \filldraw[onlineqa,draw=onlineqa,line width=.35pt] (4.69,5.76) circle[radius=.90pt];
  \node[anchor=west,font=\rmfamily\scriptsize,text=black!78] at (5.01,5.76)
    {Co-trained QA (Qt)};

  \node[anchor=south,font=\rmfamily\scriptsize\bfseries,text=black!72] at (1.75,4.65) {55.8};
  \foreach \x/\y/\value in {2.92/3.144/49.2,4.09/3.456/50.8,5.26/2.566/46.4,6.43/2.566/46.4,7.60/2.121/44.3} {
    \node[anchor=north,font=\rmfamily\scriptsize\bfseries,text=frozenqa] at (\x,\y-.12) {\value};
  }
  \foreach \x/\y/\value in {2.92/4.229/54.5,4.09/4.278/54.8,5.26/4.130/54.1,6.43/3.323/50.1,7.60/4.694/56.8} {
    \node[anchor=south,font=\rmfamily\scriptsize\bfseries,text=onlineqa] at (\x,\y+.10) {\value};
  }
  \draw[gray!62,<->,>=Latex,line width=.55pt] (7.78,2.121) -- (7.78,4.694);
  \node[fill=white,inner sep=1.2pt,font=\rmfamily\scriptsize,text=black!70]
    at (7.18,3.44) {+12.5 pp};

  \draw[gray!38,line width=.42pt] (8.35,.84) -- (8.35,6.02);
  \foreach \y/\ticklabel in {1.25/0,2.215/20,3.179/40,4.144/60,5.109/80} {
    \draw[gray!22,densely dashed,line width=.30pt] (9.25,\y) -- (17.20,\y);
    \node[anchor=east,font=\rmfamily\scriptsize,text=black!68] at (9.08,\y) {\ticklabel};
  }
  \draw[black!76,line width=.62pt] (9.25,1.25) -- (17.20,1.25);
  \draw[black!76,line width=.62pt] (9.25,1.25) -- (9.25,5.35);
  \node[rotate=90,font=\rmfamily\small,text=black!84] at (8.52,3.30)
    {Accuracy (\%)};
  \node[font=\rmfamily\small,text=black!84] at (13.23,.39) {Evaluation benchmark};

  \path[fill=qwenbar] (9.34,5.80) rectangle (9.54,5.95);
  \node[anchor=west,font=\rmfamily\tiny,text=black!78] at (9.61,5.875)
    {Qwen~\citep{qwen3technicalreport}};
  \path[fill=mmdrbar] (11.98,5.80) rectangle (12.18,5.95);
  \node[anchor=west,font=\rmfamily\tiny,text=black!78] at (12.25,5.875)
    {MM-DR~\citep{yao2026mm}};
  \path[fill=openbar] (14.35,5.80) rectangle (14.55,5.95);
  \node[anchor=west,font=\rmfamily\tiny,text=black!78] at (14.62,5.875)
    {OpenSearch~\citep{chen2026opensearch}};
  \path[fill=odebar] (9.34,5.54) rectangle (9.54,5.69);
  \node[anchor=west,font=\rmfamily\tiny,text=black!78] at (9.61,5.615)
    {ODE~\citep{huang2026towards}};
  \path[fill=visionbar] (11.98,5.54) rectangle (12.18,5.69);
  \node[anchor=west,font=\rmfamily\tiny,text=black!78] at (12.25,5.615)
    {Vision-DR~\citep{huang2026vision}};
  \path[fill=mtabar] (9.34,5.28) rectangle (9.54,5.43);
  \node[anchor=west,font=\rmfamily\tiny,text=black!78] at (9.61,5.355)
    {MTA-DS~\citep{peng2026mta}};
  \path[fill=onlineqa] (12.15,5.28) rectangle (12.35,5.43);
  \node[anchor=west,font=\rmfamily\tiny,text=black!78] at (12.42,5.355)
    {\methodname{}};

  \path[fill=qwenbar]   (9.55,1.25) rectangle (9.71,2.137);
  \path[fill=odebar]    (9.71,1.25) rectangle (9.87,2.451);
  \path[fill=visionbar] (9.87,1.25) rectangle (10.03,2.234);
  \path[fill=mtabar]    (10.03,1.25) rectangle (10.19,2.542);
  \path[fill=onlineqa]  (10.19,1.25) rectangle (10.35,2.692);
  \node[anchor=south,font=\rmfamily\tiny\bfseries,text=onlineqa] at (10.27,2.77) {29.9};

  \path[fill=qwenbar]   (11.55,1.25) rectangle (11.71,2.981);
  \path[fill=mmdrbar]   (11.74,1.25) rectangle (11.90,3.078);
  \path[fill=openbar]   (11.93,1.25) rectangle (12.09,3.064);
  \path[fill=odebar]    (12.12,1.25) rectangle (12.28,3.271);
  \path[fill=visionbar] (12.31,1.25) rectangle (12.47,3.305);
  \path[fill=mtabar]    (12.50,1.25) rectangle (12.66,3.389);
  \path[fill=onlineqa]  (12.69,1.25) rectangle (12.85,3.584);
  \node[anchor=south,font=\rmfamily\tiny\bfseries,text=onlineqa] at (12.77,3.66) {48.4};

  \path[fill=qwenbar]   (13.55,1.25) rectangle (13.71,4.003);
  \path[fill=mmdrbar]   (13.74,1.25) rectangle (13.90,4.521);
  \path[fill=openbar]   (13.93,1.25) rectangle (14.09,4.361);
  \path[fill=odebar]    (14.12,1.25) rectangle (14.28,4.434);
  \path[fill=visionbar] (14.31,1.25) rectangle (14.47,4.607);
  \path[fill=mtabar]    (14.50,1.25) rectangle (14.66,5.080);
  \path[fill=onlineqa]  (14.69,1.25) rectangle (14.85,5.022);
  \node[anchor=south,font=\rmfamily\tiny\bfseries,text=onlineqa] at (14.77,5.08) {78.2};

  \path[fill=qwenbar]   (15.55,1.25) rectangle (15.71,4.348);
  \path[fill=mmdrbar]   (15.74,1.25) rectangle (15.90,4.588);
  \path[fill=openbar]   (15.93,1.25) rectangle (16.09,4.699);
  \path[fill=odebar]    (16.12,1.25) rectangle (16.28,4.371);
  \path[fill=visionbar] (16.31,1.25) rectangle (16.47,4.371);
  \path[fill=mtabar]    (16.50,1.25) rectangle (16.66,4.774);
  \path[fill=onlineqa]  (16.69,1.25) rectangle (16.85,4.925);
  \node[anchor=south,font=\rmfamily\tiny\bfseries,text=onlineqa] at (16.77,5.00) {76.2};

  \foreach \x/\ticklabel in {10.01/MMSrch+,12.20/BC-VL,14.20/MMSrch,16.20/FVQA} {
    \draw[black!76,line width=.55pt] (\x,1.25) -- (\x,1.12);
    \node[anchor=north,font=\rmfamily\scriptsize,text=black!78] at (\x,1.05) {\ticklabel};
  }
\end{tikzpicture}}
  \vspace{-2mm}
  \caption{\textbf{Why co-training the QA policy matters.} \textbf{Left:} a
  frozen but valid QA policy produces progressively fewer informative questions
  as the Solver improves: the fraction of held-out QA groups with non-degenerate
  rollout outcomes ($0<p_i<1$) falls from $55.8\%$ at $S_0$ to $44.3\%$ at
  $S_{50}$. When the QA policy is updated online alongside the Solver, matched
  checkpoints retain or restore this training signal. \textbf{Right:} per-benchmark accuracy for selected $8$B
  agents on MMSrch-Plus, BC-VL, MMSrch, and FVQA. \methodname{} is highlighted
  in teal; the MMSrch bar includes the stronger reported specialist baseline.
  Missing bars denote benchmarks not reported by the corresponding
  prior work.}
  \vspace{-5mm}
  \label{fig:qa-curriculum}
\end{figure*}

This mismatch suggests an alternative: allowing training questions to
evolve with the Solver itself. 
Self-improvement and Generator--Solver
co-evolution have shown that adaptive task generation can create curricula
that track an agent's changing capabilities, in both text-based search and
vision-language reasoning
\citep{lu2026search,yue2026dr,he2025visplay,thawakar2025evolmm,wang2026v}.
However, this idea is substantially harder in multimodal deep search. 
A useful question must be grounded in the seed image, supported by external evidence, answerable through an independent multi-hop tool trajectory, and free of ambiguous answers or evidence leakage. Moreover, optimizing a QA Generator only for Solver failures encourages unanswerable questions, whereas optimizing only for success collapses the curriculum to trivial ones. 
The challenge is therefore to train a QA Generator that produces verified questions at the evolving Solver's learning frontier. 
Figure~\ref{fig:qa-curriculum} previews
this curriculum mismatch and the effect of adapting the QA policy online.

We address this challenge with \methodname{}, which adapts MTA-Agent's
verified QA-construction pipeline~\citep{peng2026mta} for asymmetric online
co-training of two separate models: a QA Generator and a Solver. The
QA Generator is trained to produce appropriately challenging questions for the
Solver, while the Solver is trained to answer multimodal deep-search questions.
Given a seed image and QA pair, the QA Generator 
proposes multi-hop multimodal deep search QA pairs,
and the Solver receives only
the image and final question, then produces multiple independent tool-use
rollouts.
For each generated question, the fraction of correct Solver rollouts determines both learning signals: the QA Generator is rewarded for mixed outcomes, which indicate questions that are challenging but solvable, while the Solver is updated with GRPO on the same rollout groups. As the Solver improves, its rollout feedback trains the QA Generator to target the Solver's remaining capability gaps. The two models use separate parameters, optimizers, and gradient updates.

\textbf{Contributions.}
We propose \methodname{}, an asymmetric online QA--Solver co-training
framework that updates both a \textbf{m}ultimodal \textbf{d}eep-\textbf{s}earch
QA Generator and a Solver online. The QA Generator is trained from Solver rollout
feedback to generate appropriately challenging questions, while the Solver is
trained on the resulting question groups with mixed rollout outcomes.
\methodname{} improves an $8$B Solver from $36.91\%$ to $51.58\%$ mean
accuracy across six evaluation benchmarks, outperforming the compared
$7$--$8$B specialist agents on four of six benchmarks and approaching GPT-$5$
agent performance with an $8$B model. The same pipeline also transfers to a
newer $27$B backbone. It requires no human intervention or manual filtering, no
pre-generated multi-hop corpus, and no additional task-specific SFT stage.

\vspace{-1mm}
\section{Related Work}
\vspace{-1mm}
\subsection{Multimodal Deep Search and Data Construction}
\vspace{-1mm}
Multimodal deep-search agents combine visual reasoning with external tools
such as text and image search, browsing, and OCR
\citep{wu2025mmsearch,narayan2025deepmmsearch,geng2026webwatcher,chng2025sensenova}.
Their long-horizon supervision is commonly prepared offline. A typical
pipeline first selects seed images or questions, uses capable models and tools
to synthesize candidate multi-hop questions, answers, and search trajectories,
then verifies evidence support and filters invalid, ambiguous, duplicate, or
shortcut-prone examples before assembling SFT or RL corpora. MTA-Agent
\citep{peng2026mta}, DeepMMSearch-R1 \citep{narayan2025deepmmsearch},
VSearcher \citep{zhang2026vsearcher}, MM-DeepResearch \citep{yao2026mm},
and OpenSearch-VL \citep{chen2026opensearch} instantiate variants of this
recipe with tool-assisted generation and quality, difficulty, or trajectory
filtering. REDSearcher \citep{chu2026redsearcher} and HyperEyes
\citep{li2026hypereyes} further combine such offline curation with
controllable task complexity and efficiency-oriented trajectory selection,
respectively.
ODE \citep{huang2026towards} makes the pipeline more policy-aware through on-policy data evolution. 
In contrast, we jointly train an image-grounded QA generator and a multi-tool
Solver online. The Solver's group-wise success rate
provides a Goldilocks signal that updates the QA generator; as the Solver
improves, the generator learns to pose new questions at its shifting learning
frontier, and the Solver is in turn trained on these online-generated
questions. Unlike a fixed, potentially stronger offline generator that
prepares a broadly useful corpus for many downstream models, our QA generator
is optimized against the current Solver and can focus on questions that expose
its specific capability gaps. Thus, the training distribution adapts online
rather than being fixed in a pre-generated multi-hop corpus.

\vspace{-1mm}
\subsection{Self-Improving and Co-Evolution}
\vspace{-1mm}

In text-based search, Search Self-play \citep{lu2026search} and Dr.~Zero
\citep{yue2026dr} co-evolve task proposal and solving to create an adaptive
curriculum. R-Zero \citep{huang2026r}, Absolute Zero
\citep{zhao2026absolute}, Tool-R0 \citep{acikgoz2026tool}, and Agent0
\citep{xia2025agent0} extend this principle to reasoning and general tool-use
environments. For vision-language models, VisPlay \citep{he2025visplay},
EvoLMM \citep{thawakar2025evolmm}, and V-Zero \citep{wang2026v} use
Questioner--Solver loops over visual inputs, while Vision-Zero
\citep{wang2025vision}, iReasoner \citep{sunil2026ireasoner}, and Agent0-VL
\citep{liu2025agent0} explore strategic self-play, intrinsic rewards, or
tool-grounded self-refinement.
Our setting instead targets open-world multimodal deep search.
Although \methodname{} follows the same co-evolutionary intuition, multimodal deep search poses a substantially harder task-generation problem: the QA Generator must retrieve and verify external evidence, preserve visual grounding and entity continuity across multiple hops, and calibrate question difficulty for an independently acting Solver that uses heterogeneous tools. A question that appears plausible from an image may still be unanswerable, ambiguous, or unsupported by the retrieved evidence, making naive task proposal unreliable. \methodname{} therefore combines a fixed verification harness with Solver-calibrated rewards to construct a reliable, capability-adaptive curriculum.

\vspace{-1mm}
\section{Method}
\vspace{-1mm}
\label{sec:method}

\begin{figure*}[tb!]
  \centering
  \includegraphics[width=\textwidth]{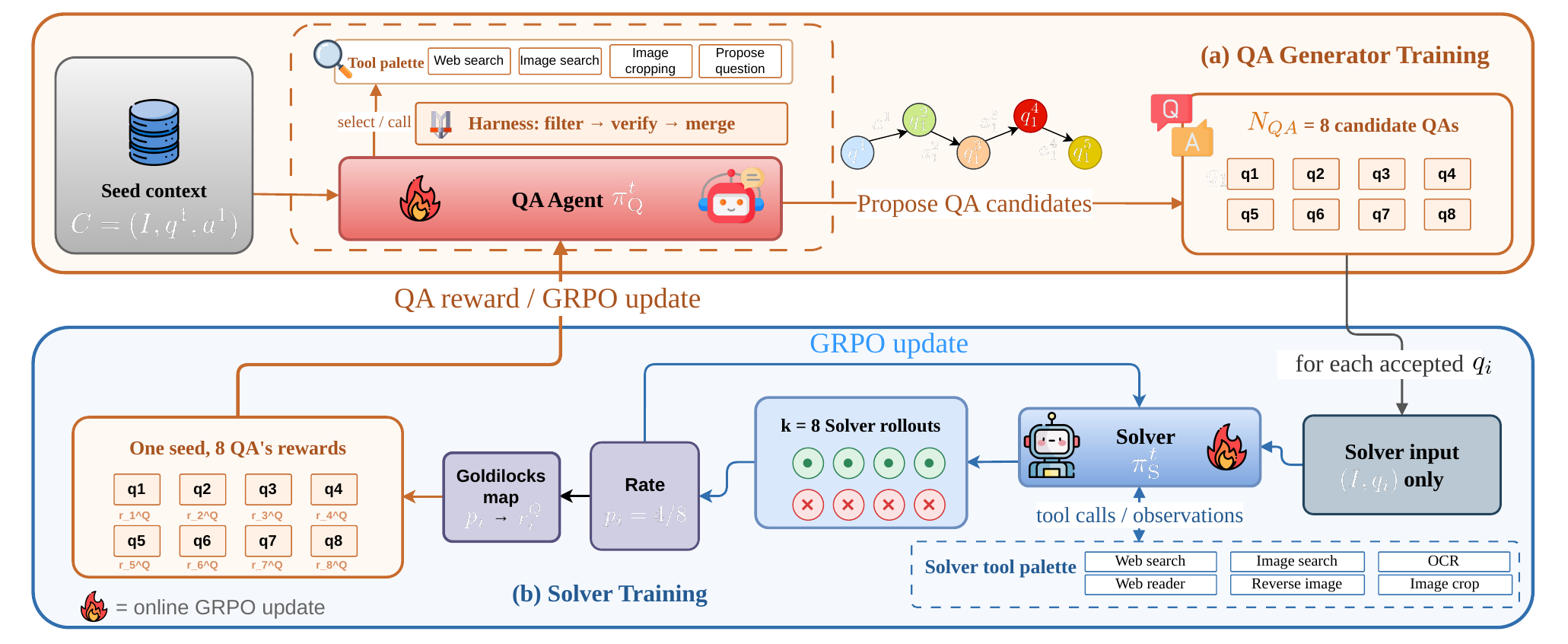}
  \vspace{-5mm}
  \caption{\textbf{Overview of \methodname.}
  The QA Generator and Solver are updated online from separate GRPO objectives,
  while the fixed harness executes and validates QA-construction actions. The
  QA-side tool palette depicts the general harness interface; the main
  experiments enable web-text search only for QA construction.
  }
  \label{fig:mds-zero-overview}
  \vspace{-5mm}
\end{figure*}

As Figure~\ref{fig:mds-zero-overview} shows, \methodname{} co-trains a training-time QA Generator \(\pi_{\mathrm{Q}}\)
and a target multimodal deep-search Solver \(\pi_{\mathrm{S}}\). At
iteration \(t\), we start from an image-grounded seed context
\(C=(I,q^{1},a^{1})\). 
The QA generator \(\pi_{\mathrm{Q}}^t\) (\S\ref{sec:qa-generator}) produces \(N_{\mathrm{QA}}\) evidence-grounded candidate QA pairs \(\{(I, q_i,a_i)\}_{i=1}^{N_{\mathrm{QA}}}\). 
For every accepted \(q_i\), the Solver \(\pi_{\mathrm{S}}^t\) (\S\ref{sec:solver}) receives \((I,q_i)\) and makes \(K\)
independent multi-step tool-use rollouts. 
Let \(p_i\) be the fraction of those attempts
whose final answers are correct. 
This one scalar closes the loop in two distinct ways. First, we map \(p_i\) to a Goldilocks reward (\S\ref{sec:qa-reward}) for
the QA trajectory that proposed \(q_i\): questions solved by all rollouts or by none receive no QA reward, while questions near the target success rate are preferred. 
Second, the \(K\) attempts form one Solver GRPO group. 
The QA generator receives only this scalar-derived reward; the Solver is updated from its own individual rollout outcomes. 
No gradient is passed between the two policies, which retain separate parameters and optimizers.

\vspace{-2mm}
\subsection{Online QA Generation with a Harness}
\vspace{-2mm}
\label{sec:qa-generator}

The QA Generator is a trainable agent that constructs challenging multi-hop multimodal QA
pairs for training the deep-search Solver agent. It includes:
($1$) a trainable LLM policy that writes web-search queries, then
proposes a single-hop question--answer pair grounded in the retrieved evidence;
and ($2$) a fixed, non-trainable harness that executes the policy's actions,
checks the validity of each proposed hop, and merges accepted single-hop QA pairs into a final multi-hop QA task.

\textbf{Growing an \(n\)-hop question.}
Rather than precomputing a generic training corpus, the QA Generator
\(\pi_{\mathrm{Q}}\) converts a seed context \(C=(I,q^{1},a^{1})\) into a
multi-hop QA pair online.  We count $C$ as the first
of \(n\) hops and set \(\bar q^{1}:=q^{1}\); its answer \(a^{1}\) is the
initial entity to explore.
For each subsequent hop
\(h\in\{2,\ldots,n\}\),
\(\pi_{\mathrm{Q}}\) searches for information about the previous answer
\(a^{h-1}\), then proposes an evidence-based QA pair \((q^{h},a^{h})\).  The new
question \(q^{h}\) asks for a fact about the previous entity \(a^{h-1}\), while its answer
\(a^{h}\) identifies the entity used to extend the next hop. For each non-final
hop, this entity must be specific, and evidence-supported (Appendix~\ref{app:entity-state}). Once the proposal
is accepted, the harness combines \(q^{h}\) with the running question
\(\bar q^{h-1}\) to produce \(\bar q^{h}\).  Crucially, this merge removes the explicit mention of
\(a^{h-1}\): the resulting question reaches \(a^{h}\) through the preceding
description rather than revealing the intermediate answer.  Repeating this
process to the target depth \(n\) yields
\((I,q_i,a_i)=(I,\bar q^{n},a^{n})\). 
An example is illustrated in Appendix~\ref{app:qa-example}.

\textbf{Learned LLM policy.}
The learned policy selects the next action
and writes that action's arguments. It alternates between ($1$) a
\textit{web-search} action containing a query and ($2$) a
\textit{propose-next-hop} action containing a candidate question, answer,
supporting evidence, and a source field.
Thus, the
policy learns both what information to seek and how to turn that information
into the next QA hop.  

\textbf{Non-trainable QA harness.}
Our fixed harness adapts the offline verification-and-merging pipeline of
MTA-Agent~\citep{peng2026mta} for online policy training.
After a search action, it retrieves pages, filters them into bounded evidence chunks, and
returns those chunks as an observation for the subsequent proposal action.
After a proposed hop, it checks that the
question is anchored to, and actually requires, the current entity.  
It then checks the answer uniqueness, temporal stability, and self-contained wording.
It rejects source-referential,
vague, generic, or otherwise non-chainable proposals. 
Finally, the harness merges an
accepted proposal with the running question, verifies that the merged question
uniquely resolves to the intended answer, and discards questions that a fixed
direct-answer check can solve without search.
More details can be found in the Appendix~\ref{app:qa_harness}.

\textbf{Harness feedback and accepted-hop training.}
For every proposed hop, the harness appends a feedback message to the context as
a user turn. 
When a hop is accepted, this message contains the updated merged
question \(\bar q^h\), the next entity \(a^h\), and the acceptance signal; the
QA Generator then uses this updated context to select its next action.
Only action trajectories that lead to an accepted hop are used to train
\(\pi_{\mathrm{Q}}\). 
If a proposal is rejected, its
retrieved evidence and proposed hop are retained only as context and diagnostic
feedback for a new attempt; all policy tokens associated with that unsuccessful
attempt are masked from the training loss.
Representative feedback and token-level credit assignment are provided in
Appendix~\ref{app:qa-feedback-examples} and
Appendix~\ref{app:harness-boundary}, respectively.

\vspace{-3mm}
\subsection{Multi-tool Solver Rollouts}
\vspace{-1mm}
\label{sec:solver}

For each accepted training question \(q_i\), we sample \(K\) independent ReAct rollouts. 
At step \(t\) of rollout \(j\), the Solver either  selects a tool and generates its arguments or terminates by producing a final answer. 
When a tool is selected, the environment executes the call and appends the returned observation to the history, yielding \(H_{ij}^{t+1}\) for the next decision. This interaction loop continues until the Solver returns a final answer or reaches the maximum number of interaction steps.
Let \(c_{ij}\in\{0,1\}\) denote the terminal correctness of rollout \(j\) for
question \(q_i\), determined by answer matching (Appendix~\ref{app:llm-judge}). 
The group success rate is
\begin{equation}
  p_i = \frac{1}{K}\sum_{j=1}^{K} c_{ij}.
  \label{eq:solver-rate}
\end{equation}
We update \(\pi_{\mathrm{S}}\) with standard GRPO only when \(0<p_i<1\).
These groups contain both correct and incorrect rollouts and therefore provide
a useful relative learning signal.

\vspace{-1mm}
\subsection{Solver-Calibrated QA Scores}
\vspace{-1mm}
\label{sec:qa-reward}

The same rate \(p_i\) supplies an adaptive difficulty signal for the QA
generator. We map it to a Goldilocks score,
\begin{equation}
 r^{\mathrm{Q}}_i =
 \begin{cases}
  0, & p_i \in \{0,1\},\\
  \dfrac{p_i}{\tau}, & 0 < p_i \leq \tau,\\
  \dfrac{1-p_i}{1-\tau}, & \tau < p_i < 1,
 \end{cases}
 \label{eq:qa-reward}
\end{equation}
where \(\tau\in(0,1)\) is the target Solver success rate. The reward peaks at
\(p_i=\tau\), while questions solved by every rollout or by none receive zero
reward. Optimizing \(\pi_{\mathrm{Q}}\) with GRPO therefore favors questions
that are difficult but still solvable by the current Solver, rather than
questions that merely maximize failure.
For clarity, our main setting uses \(K=8\) and \(\tau=0.5\). Thus, when
\(k=0,\ldots,8\) of the eight Solver rollouts answer a question correctly, its
QA score is
\[
\begin{array}{c|ccccccccc}
k & 0 & 1 & 2 & 3 & 4 & 5 & 6 & 7 & 8 \\
\hline
r_i^{\mathrm{Q}} & 0 & .25 & .50 & .75 & 1.00 & .75 & .50 & .25 & 0
\end{array}.
\]

Every QA candidate is scored separately. For one seed, the figure's
\(N_{\mathrm{QA}}=8\) candidates yield the reward set
\(\{r_i^{\mathrm{Q}}\}_{i=1}^{N_{\mathrm{QA}}}\), which is used to optimize
\(\pi_{\mathrm{Q}}\) with GRPO over these QA-generation trajectories. 
This per-candidate assignment is what
allows the curriculum to track the Solver's changing learning frontier.




\vspace{-1mm}
\section{Experiments}
\vspace{-1mm}
\label{sec:experiments}


\vspace{-1mm}
\subsection{Experiment Setup}
\vspace{-1mm}
\label{sec:experimental-setup}

\subsubsection{Datasets}
\vspace{-1mm}
\label{sec:evaluation-datasets}

\textbf{Evaluation.}
We evaluate the trained Solver on six benchmarks that require visual evidence
understanding and open-web research: MMSearch-Plus (MMSrch-Plus)~\citep{tao2025mmsearch},
HR-MMSearch-VL (HR-MMSrch)~\citep{chng2025sensenova}, BrowseComp-VL
(BC-VL)~\citep{geng2025webwatcher}, MMSearch-VL (MMSrch)~\citep{jiang2024mmsearch},
the testmini split of VDR-Bench (VDR)~\citep{huang2026vision}, and the
official FVQA test split (FVQA)~\citep{wu2025mmsearch}.

\textbf{Training seeds.}
Following the publicly described seed-source protocol of MTA-Agent
\citep{peng2026mta}, we initialize co-training from a fixed image-grounded
one-hop VQA pool: InfoSeek~\citep{chen2023can},
the training split of FVQA~\citep{wu2025mmsearch},
InfoVQA~\citep{mathew2022infographicvqa}, and 
OK-VQA~\citep{okvqa}.
Each seed contains only
\(C=(I,q^1,a^1)\): an image, its original one-hop question, and its answer.
Appendix~\ref{app:seed-selection} gives the exact pool
composition and filtering rules.
The pool provides initial visual grounding rather than an offline deep-search
corpus: it contains no pre-generated multi-hop QA pairs, retrieved evidence,
or search trajectories. All multi-hop questions used for optimization are
subsequently constructed and verified online by the QA generator and harness in Section~\ref{sec:qa-generator}.

\vspace{-1mm}
\subsubsection{Agent Setup}
\vspace{-1mm}
\label{sec:solver-setup}

\textbf{Solver Setup.}
We use a \textbf{ReAct} (Reasoning and Acting) framework~\citep{yao2022react}
for the deep-search Solver (\S\ref{sec:solver}).
It operates for at most $T=10$ iterations. At each iteration, it generates a structured output comprising (i) a \textit{reasoning} trace based on the current state, (ii) an \textit{action} specifying a tool and its parameters, (iii) a binary \texttt{should\_stop} flag, and (iv) a scalar \textit{confidence} score. The loop terminates early when \texttt{should\_stop=true} and the confidence score exceeds $0.7$, or the maximum budget of $T=10$ iterations is exhausted. Upon termination, the agent generates its final answer conditioned on the complete interaction history. Further details are provided in Appendix~\ref{app:solver-interaction}.

\textbf{QA Agent setup.}
The QA Generator follows the construction-specific interaction loop described in Section~\ref{sec:qa-generator}. For each seed context, we sample $8$ QA-generation trajectories with \texttt{max\_turns} set to $16$. At a search step, the Generator emits \texttt{action}, \texttt{thought}, \texttt{tool}, and \texttt{query}; the harness executes web-text search, filters the returned pages into evidence chunks, and appends them to the next interaction state. At a proposal step, it emits \texttt{action}, \texttt{thought}, \texttt{question}, \texttt{answer}, \texttt{evidence}, and \texttt{source}. The harness verifies the proposal and, if accepted, appends the merged question and next entity; otherwise, it returns diagnostic feedback. A trajectory terminates after reaching the target depth or exhausting its configured turn budget. Appendix~\ref{app:entity-state} specifies the state and action schemas.

\textbf{Tool Palette.}
The Solver has access to: \textit{Web search} retrieves web
results, including snippets, titles, and URLs. \textit{Image search}
returns textual descriptions of images found on the web. \textit{Web URL
reader} extracts the full text of a given webpage. \textit{Reverse image
search} uses Google Lens to identify objects and landmarks. \textit{OCR} extracts visible text from an image, and
\textit{image cropping} extracts a specified region from an image.  All Solver
tool calls are sent to a local observation server, and the returned observation
is appended to the chat history for the next action.  More details are in Appendix~\ref{app:tools}.

\vspace{-2mm}
\subsubsection{Baselines}
\vspace{-1mm}

We compare to three commercial large language models: GPT-$5$~\citep{singh2025openai}, Claude Sonnet $4.5$, and Gemini $3$ Pro~\citep{gemini3_2025}. We evaluate each model under two settings: ($1$) direct answering without tools and ($2$) agent-based answering using the same framework and tool configuration as our method. 
In addition, we report results from prior work: MTA-DeepSearch~\citep{peng2026mta}, MMSearch-R1~\citep{wu2025mmsearch}, SenseNova-MARS~\citep{chng2025sensenova}, WebWatcher~\citep{geng2025webwatcher}, VSearcher~\citep{zhang2026vsearcher}, MM-DeepResearch~\citep{yao2026mm}, OpenSearch-VL~\citep{chen2026opensearch}, ODE~\citep{huang2026towards}, and Vision-DeepResearch.

\vspace{-1mm}
\subsubsection{Training Setup}
\vspace{-1mm}

The QA generator and Solver begin from separate copies of
\texttt{Qwen3-VL-8B-Instruct}~\citep{qwen3technicalreport}. Instead of
synthesizing a multi-hop corpus before training, we initialize online
generation from a fixed pool of image-grounded, one-hop VQA seed
contexts in Section~\ref{sec:evaluation-datasets}. 
Each optimization iteration draws $8$ seed QAs from the pool. 
The QA policy generates $8$ trajectories targeting a total depth of $5$ hops
(one seed hop and four extensions) per seed, and every retained QA is answered by
$8$ independently sampled Solver rollouts. 
This produces $64$ QA trajectories
and $512 = 8 \times 8 \times 8$ Solver trajectories in one iteration. 
We optimize the two policies independently with GRPO for $50$ iterations, with learning rate
\(10^{-6}\), and Goldilocks target \(\tau=0.5\). 
The QA generator and Solver have turn caps of $16$ and $10$; their prompt and response limits are
$8,192$ and $24,576$ tokens, respectively. 
All main $8$B numbers come from the Solver after the $50$th co-training iteration.
Full configuration is in Appendix~\ref{app:training-details} and online data and
resource accounting are in Appendix~\ref{app:data-compute}.

\vspace{-2mm}
\subsection{Main Results}
\vspace{-2mm}
\label{sec:main-results}

\textbf{Strong improvement without a pre-generated multi-hop corpus.}
As shown in Table~\ref{tab:main-results}, 
comparing to the matched
\texttt{Qwen3-VL-8B-Instruct} Agent Workflow baseline (in gray), \methodname{} raises
the six benchmark mean accuracy from $36.91\%$
to $51.58\%$ ($+14.67$ points).
The run uses $400$ one-hop seed-context draws over $50$ co-training iterations: the
QA generator and Solver are optimized directly with GRPO, without a separate
multi-hop data factory or an additional task-specific SFT stage. Full resource
and service-use accounting is reported in Appendix~\ref{app:data-compute}.


\textbf{Comparison with $7$--$8$B specialist deep-search agents.}
Among the specialist agents listed in Table~\ref{tab:main-results},
\methodname{} records the highest score on four of six benchmarks.
Prior specialist agents including MTA-Agent~\citep{peng2026mta}, VSearcher~\citep{zhang2026vsearcher}, and OpenSearch-VL~\citep{chen2026opensearch} first construct, filter, or reuse offline teacher-assisted QA or trajectory corpora before applying SFT and/or RL. 
Because their results use their original tool stacks and training pipelines, this is not a cost-matched comparison. 
However, \methodname{} and MTA-DeepSearch~\citep{peng2026mta} share the same raw one-hop seed pool: \methodname{} replaces MTA-DeepSearch's offline multi-hop data-construction pipeline with online QA--Solver co-training. Without pre-generating a multi-hop corpus or adding an additional task-specific SFT stage, \methodname{} outperforms MTA-DeepSearch on four of the five benchmarks for which both methods report results.

\textbf{Approaching GPT-5 at the 8B scale.}
Despite using an $8$B open model, \methodname{} reaches $51.58\%$ average
accuracy across six benchmarks, coming within $0.40$ percentage points of
GPT-$5$ under the same Agent Workflow ($51.98\%$). Although stronger
commercial agents remain ahead overall, this result shows that online
QA--Solver co-training can make a compact model competitive with a strong
frontier system without increasing the Solver backbone size.

\begin{table*}[t]
    \centering
    \vspace{-2mm}
    \small
    \caption{Comparison on multimodal deep-search benchmarks. Results report mean accuracy (\%) over 3 generation seeds. Avg is the unweighted mean across all six benchmarks and is
    reported only when all six scores are available. \(\dagger\) marks
    results reported by the original papers.
    }
    \setlength{\tabcolsep}{5pt}
    \resizebox{\textwidth}{!}{%
    \begin{tabular}{l|rrrrrr|r}
    \toprule
    Model & MMSrch-Plus & HR-MMSrch & BC-VL & MMSrch & FVQA & VDR & Avg \\
    \midrule
    \multicolumn{8}{c}{\textbf{Direct Answer}} \\
    \midrule
    GPT-5                 & 13.18 & 16.39 & 41.60 & 36.47 & 50.83 & 14.40 & 28.81 \\
    Claude 4.5 Sonnet     &  5.40 & 14.40 & 36.10 & 40.00 & 49.60 & 7.00 & 25.42 \\
    Gemini-3-Pro          & 26.37 & 23.61 & 41.35 & 65.88 & 58.94 & 14.00 & 38.36 \\
    Qwen3-VL-8B-Instruct  &  1.29 &  4.59 & 20.05 & 12.35 & 26.94 & 3.00 & 11.37 \\
    \midrule
    \multicolumn{8}{c}{\textbf{Agent Workflow}} \\
    \midrule
    GPT-5                 & 31.61 & 52.13 & 51.63 & 77.65 & 72.28 & 26.60 & 51.98 \\
    Claude 4.5 Sonnet     & 43.00 & 53.40 & 65.70 & 84.70 & 77.00    & 24.40 & 58.03 \\
    Gemini-3-Pro          & 33.51 & 53.20 & 51.78 & 82.94 & 76.67 & 27.60 & 54.28 \\
    \rowcolor{gray!15} Qwen3-VL-8B-Instruct  & 18.38 & 32.13 & 35.89 & 57.06 & 64.22 & 13.80 & 36.91 \\
    \midrule
    \midrule
    \multicolumn{8}{c}{\textbf{Multimodal Deep-Search Agents}} \\
    \midrule
    MMSearch-R1-7B         & -- & -- & -- & 53.80$^{\dagger}$ & 58.40$^{\dagger}$ & -- & -- \\
    SenseNova-MARS-8B      & -- & 41.64$^{\dagger}$ & -- & 67.84$^{\dagger}$ & 67.11$^{\dagger}$ & -- & -- \\
    WebWatcher-7B          & -- & -- & 21.20$^{\dagger}$ & 49.10$^{\dagger}$ & -- & -- & -- \\
    VSearcher-8B           & -- & -- & 30.80$^{\dagger}$ & 47.20$^{\dagger}$ & -- & -- & -- \\
    MM-DeepResearch-8B     & -- & -- & 37.90$^{\dagger}$ & 67.80$^{\dagger}$ & 69.20$^{\dagger}$ & -- & -- \\
    OpenSearch-VL-8B       & -- & -- & 37.60$^{\dagger}$ & 64.50$^{\dagger}$ & 71.50$^{\dagger}$ & 20.80$^{\dagger}$ & -- \\
    ODE-8B-RL              & 24.90$^{\dagger}$ & -- & 41.90$^{\dagger}$ & 66.00$^{\dagger}$ & 64.70$^{\dagger}$ & 20.40$^{\dagger}$ & -- \\
    Vision-DeepResearch-8B & 20.40$^{\dagger}$ & -- & 42.60$^{\dagger}$ & 69.60$^{\dagger}$ & 64.70$^{\dagger}$ & \textbf{29.20}$^{\dagger}$ & -- \\
    MTA-DeepSearch-8B      & 26.77$^{\dagger}$ & 47.54$^{\dagger}$ & 44.36$^{\dagger}$ & \textbf{79.41}$^{\dagger}$ & 73.06$^{\dagger}$ & -- & -- \\
    \midrule
    \rowcolor{green!15}\methodname{} (Qwen3-VL-8B) & \textbf{29.90} & \textbf{53.10} & \textbf{48.40} & 78.20 & \textbf{76.20} & 23.65 & 51.58 \\
    \bottomrule
    \end{tabular}%
    }
    \label{tab:main-results}
    \end{table*}

\section{Analysis}
\label{sec:analysis}

\subsection{Does the QA Agent Learn an Adaptive Curriculum?}
\vspace{-1mm}
\label{sec:qa-curriculum-analysis}

Before online co-training, the pretrained base model and our fixed QA harness already function as a QA agent: they can search evidence, propose next-hop QA pairs, and generate verified multi-hop questions. The key question is \textit{whether further training of this agent is necessary}. 
We use checkpoint cross-play to assess it. 
Starting from $128$ held-out seed QA pairs, the agent generates $4$ questions $q_i$ per seed. For each generated question, we estimate the Solver’s success rate $p_i$ using $8$ independent rollouts. We define the \emph{informative group rate} (I.G. rate) as the proportion of questions with $0 < p_i < 1$, because only these questions yield non-degenerate GRPO groups for the Solver.

The left panel of Figure~\ref{fig:qa-curriculum} tests whether the initial QA agent, $Q_0$ (\texttt{Qwen3-VL-8B-Instruct}), remains suitable as the Solver evolves. Its dashed red curve fixes $Q_0$ and varies the Solver from $S_0$ to $S_{50}$. The I.G. rate falls from $55.8\%$ to $44.3\%$. 
Thus, even a capable frozen QA agent and fixed harness become mismatched to a stronger Solver.
We then fix the Solver at $S_{50}$ and use successive QA checkpoints to generate $4$ questions from each of the same $128$ held-out seed QA pairs ($512$ questions in total). The right panel of Table~\ref{tab:qa-curriculum-crossplay} shows that the I.G. rate rises from $44.3\%$ for $Q_0$ to $56.8\%$ for $Q_{50}$, a gain of $12.5$ percentage points. Since the Solver, seed contexts, and fixed harness are unchanged, this gain comes from the learned QA policy.

The left plot of Table~\ref{tab:qa-curriculum-crossplay} reveals how the distribution changes. Relative to $Q_0$, $Q_{50}$ reduces universally failed groups ($p_i=0$) from $27.5\%$ to $18.6\%$ and universally solved groups ($p_i=1$) from $28.1\%$ to $24.6\%$. At the same time, it raises the proportion of groups in the central $3/8$--$5/8$ success-rate range from $16.6\%$ to $24.6\%$. The solid teal curve in Figure~\ref{fig:qa-curriculum} shows the complementary matched-checkpoint view: its I.G. rate stays between $50.1\%$ and $56.8\%$ throughout co-training. Together, these results show that the QA policy learns to generate questions that remain useful for the current Solver, rather than supplying a static distribution that is useful only near initialization.

\begin{table*}[t]
    \centering
    \vspace{-2mm}
    \caption{QA--Solver checkpoint cross-play. \textbf{Left:} with the final
    Solver fixed at $S_{50}$, the learned QA policy shifts generated tasks
    away from universally failed and universally solved rollout groups toward
    the Goldilocks region. 
    \textbf{Right:} the informative-group rate increases
    across QA checkpoints with $S_{50}$ fixed.}
    \label{tab:qa-curriculum-crossplay}
    \footnotesize
    \begin{minipage}[t]{0.46\textwidth}
        \vspace{0pt}
        \centering
        \textbf{(a) Rollout-success distribution at fixed $S_{50}$}\\[-1pt]
        \begin{tikzpicture}[x=0.62cm,y=0.045cm,baseline=(current bounding box.north)]
            \foreach \y in {0,10,20,30} {
                \draw[gray!28,line width=.25pt] (-.45,\y) -- (8.55,\y);
                \node[font=\scriptsize,anchor=east,text=gray!80]
                    at (-.57,\y) {\y};
            }
            \draw[gray!75,line width=.45pt] (-.45,0) -- (8.55,0);
            \draw[gray!75,line width=.45pt] (-.45,0) -- (-.45,32);
            \foreach \x/\qzero/\qfifty in {
                0/27.54/18.55, 1/6.45/9.18, 2/6.45/5.27,
                3/5.27/7.62, 4/5.47/8.20, 5/5.86/8.79,
                6/5.47/8.01, 7/9.38/9.77, 8/28.13/24.61}
            {
                \filldraw[fill=gray!58,draw=gray!75,line width=.25pt]
                    (\x-.28,0) rectangle (\x,\qzero);
                \filldraw[fill=teal!72!black,draw=teal!85!black,line width=.25pt]
                    (\x,0) rectangle (\x+.28,\qfifty);
                \node[font=\scriptsize,anchor=north] at (\x,-1.1) {$\x/8$};
            }
            \fill[gray!58] (6.04,33.2) rectangle ++(.24,1.8);
            \node[font=\scriptsize,anchor=west] at (6.38,34.1) {$Q_0$};
            \fill[teal!72!black] (7.13,33.2) rectangle ++(.24,1.8);
            \node[font=\scriptsize,anchor=west] at (7.47,34.1) {$Q_{50}$};
            \node[font=\tiny,rotate=90,align=center] at (-1.95,15.5)
                {\shortstack{Generated QA\\proportion (\%)}};
            \node[font=\scriptsize,anchor=north] at (4,-8.0)
                {Solver success rate $p_i$};
        \end{tikzpicture}
    \end{minipage}\hfill
    \begin{minipage}[t]{0.46\textwidth}
        \vspace{0pt}
        \centering
        \textbf{(b) QA policy improves with fixed Solver}\\[-1pt]
        \vspace{1.0mm}
        {\scriptsize Fixed Solver: $S_{50}$; vary the QA policy.}\\[2.0mm]
        \begingroup
        \scriptsize
        \setlength{\tabcolsep}{2.6pt}
        \begin{tabular*}{\linewidth}{@{\extracolsep{\fill}}lcccccc@{}}
        \toprule
        QA checkpoint & $Q_0$ & $Q_{10}$ & $Q_{20}$ & $Q_{30}$ & $Q_{40}$ & \textcolor{teal!85!black}{$Q_{50}$} \\
        \midrule
        I.G. rate (\%)
            & $44.3$ & $46.4$ & $48.4$ & $49.8$ & $49.1$ & \textcolor{teal!85!black}{\textbf{$56.8$}} \\
        \midrule
        \textit{Net change}
            & \multicolumn{6}{c}{\textcolor{teal!85!black}{\textbf{$44.3\% \rightarrow 56.8\%$ ($+12.5$ pp)}}} \\
        \bottomrule
        \end{tabular*}
        \endgroup
    \end{minipage}
\end{table*}

\vspace{-3mm}

\newcommand{\qaRewardProfileHalf}{%
\begin{tikzpicture}[baseline=-0.55ex,x=0.17cm,y=0.34cm]
    \draw[gray!55,line width=0.25pt] (-0.06,0) -- (8.78,0);
    \foreach \x/\h in {0/0,1/.25,2/.5,3/.75,4/1,5/.75,6/.5,7/.25,8/0}
        \fill[brightcerulean!88!black] (\x,0) rectangle ++(.72,\h);
\end{tikzpicture}%
}
\newcommand{\qaRewardProfileQuarter}{%
\begin{tikzpicture}[baseline=-0.55ex,x=0.17cm,y=0.34cm]
    \draw[gray!55,line width=0.25pt] (-0.06,0) -- (8.78,0);
    \foreach \x/\h in {0/0,1/.5,2/1,3/.833,4/.667,5/.5,6/.333,7/.167,8/0}
        \fill[orange!88!black] (\x,0) rectangle ++(.72,\h);
\end{tikzpicture}%
}

\begin{table*}[t]
    \centering
    \caption{Ablating the QA difficulty target on the 8B Solver. Overall is the unweighted mean across the four benchmarks.} 
    \label{tab:qa-reward-ablation}
    \footnotesize
    \setlength{\tabcolsep}{4pt}
    \begin{tabular}{c c|rrrr|r}
    \toprule
    QA target & \shortstack{QA reward profile\\[-1pt]\scriptsize $k=0,\ldots,8$}  & MMSrch-Plus & HR-MMSrch & BC-VL & MMSrch & Overall \\
    \midrule
    \(\tau=0.50\) & \qaRewardProfileHalf  & 29.9\% & 53.1\% & 48.4\% & 78.2\% & 52.4\% \\
    \(\tau=0.25\) & \qaRewardProfileQuarter  & 25.1\% & 51.1\% & 45.4\% & 74.7\% & 49.1\% \\
    \bottomrule
    \end{tabular}
\end{table*}

\begin{table}[t]
    \centering
    \vspace{-2mm}
    \caption{Comparison of fixed and adaptive QA policies for an
    8B Solver. }
    \label{tab:frozen-qa-ablation}
    \footnotesize
    \setlength{\tabcolsep}{4pt}
    \resizebox{0.94\columnwidth}{!}{%
    \begin{tabular}{l|cccc|c}
    \toprule
    Method & MMSrch-Plus & HR-MMSrch & BC-VL & MMSrch & Avg \\
    \midrule
    \rowcolor{gray!15} Qwen3-VL-8B-Instruct (base) & 18.38 & 32.13 & 35.89 & 57.06 & 35.87 \\
    \methodname{} (frozen QA) & 24.89 & 42.30 & 42.86 & 67.06 & 44.28 \\
    \methodname{} (adaptive QA policy) & \textbf{29.90} & \textbf{53.10} & \textbf{48.40} & \textbf{78.20} & \textbf{52.40} \\
    \bottomrule
    \end{tabular}%
    }
    
\end{table}

\begin{table*}[t]
    \centering
    \small
    \caption{Large-model context for scaling \methodname{} to Qwen3.6-27B.}
    \label{tab:qwen36-27b-scaling}
    \setlength{\tabcolsep}{5.5pt}
    \resizebox{\textwidth}{!}{%
    \begin{tabular}{l|rrrr|r}
    \toprule
    Method & MMSrch-Plus & HR-MMSrch & BC-VL & MMSrch & Overall (ours) \\
    \midrule
    \multicolumn{6}{c}{\textbf{Reported 27--32B Multimodal Deep-Search Agents}} \\
    \midrule
    MTA-DeepSearch-32B$^{\dagger}$                & 31.93 & 53.95 & 53.77 & 82.35 & -- \\
    ODE-30B-A3B-RL (MoE)$^{\dagger}$              & 28.10 & --    & 46.10 & 69.70 & -- \\
    REDSearcher-MM-RL-30B (MoE)$^{\dagger}$       & 26.60 & --    & 57.20 & 72.90 & -- \\
    HyperEyes-30B-RL (MoE)$^{\dagger}$            & 31.50 & --    & 57.90 & 86.90 & -- \\
    Vision-DeepResearch-30B-A3B (MoE)$^{\dagger}$ & 28.50 & --    & 53.70 & 69.60 & -- \\
    \midrule
    \multicolumn{6}{c}{\textbf{Our Local Evaluation}} \\
    \midrule
    Qwen3.6-27B (Base)  & 36.20 & 45.90 & 46.60 & 58.80 & 46.88 \\
    \methodname{} (Qwen3.6-27B) & 38.50 & 57.00 & 54.10 & 81.20 & 57.70 \\
    \bottomrule
    \end{tabular}%
    }
\end{table*}

\subsection{Ablating the QA Difficulty Target}
\vspace{-1mm}
\label{sec:qa-reward-ablation}

The Goldilocks reward does not merely reject degenerate question groups: its
target \(\tau\) specifies the Solver success rate that the QA policy is
encouraged to produce. Our default \(\tau=0.5\) assigns maximal reward to
questions answered correctly by \(4\) of \(8\) Solver rollouts.
We also evaluate a more aggressive low-success target motivated by
Absolute Zero \citep{zhao2026absolute}. Its inverse-accuracy reward is largest
when exactly \(1\) of \(8\) Solver rollouts is correct. Rather than directly
use such a strict target, we set \(\tau=0.25\), which peaks at \(2/8\)
correct rollouts. A lone correct rollout can be a lucky sample rather than
reliable evidence that the generated question is solvable, whereas \(2\)
correct rollouts provide a more robust indication of a difficult but
repeatably solvable question.
For \(k=0,\ldots,8\) correct rollouts, this target gives the QA reward vector
\((0,.5,1,.833,.667,.5,.333,.167,0)\).

Table~\ref{tab:qa-reward-ablation} compares both targets after the same
\(50\)-step co-training budget. The lower target obtains an unweighted overall
score of \(49.1\%\), \(3.3\) points below the \(52.4\%\) obtained by
\(\tau=0.5\).
For multi-tool deep search, difficult but reliably solvable questions are more useful than questions that are merely close to
failure. With \(K=8\), a \(2/8\) target has only one lower-success outcome before a question becomes completely unsolved (\(k=0\)) and is
discarded from the Solver update. In contrast, a \(4/8\) target keeps a wider
margin of mixed-success groups and supplies both successful and unsuccessful
rollouts for relative learning.

\subsection{Does Updating the QA Policy Improve the Solver?}
\label{sec:qa-source-analysis}
\vspace{-1mm}

The cross-play evidence above measures whether a QA policy remains informative
as its Solver changes. Table~\ref{tab:frozen-qa-ablation} provides the
outcome-level complement: how much does a valid but fixed QA policy improve the
Solver, and what does updating that policy add? The untuned
\texttt{Qwen3-VL-8B-Instruct} agent begins at $35.87\%$ average accuracy
under the same agent workflow. Holding the QA policy fixed while retaining the
online harness raises this score to $44.28\%$ (\(+8.41\) points). The frozen and
adaptive runs start from the same Solver initialization and use exactly the
same seed contexts, harness, tools, rollout budget, and training schedule;
the only training difference is whether the QA policy is updated. Thus, the
initial frozen QA policy and harness are already useful---they generate verified
questions that train the Solver beyond its base capability.
Updating the QA policy throughout training raises the average further to $52.40\%$,
an additional \(+8.12\) points over the frozen-QA control and \(+16.53\)
points over the base Solver. This pattern agrees with the checkpoint
cross-play: online QA construction alone helps, but a policy that continues to
adapt to Solver outcomes supplies a substantially stronger curriculum.

\subsection{Scaling to a Larger, Newer Backbone}

\label{sec:scaling-analysis}

Our gains are not confined to the $8$B setting. We also initialize both policies
from Qwen3.6-27B~\citep{qwen3.6-27b}, a larger and newer backbone than the Qwen3-VL-8B Solver used
in our main experiments. 
On the same seed VQA contexts, the resulting
\methodname{} raises the overall OOD accuracy from $46.88\%$
for its Qwen3.6-27B base model to $57.70\%$ (\(+10.82\) points;
Table~\ref{tab:qwen36-27b-scaling}). The improvement is positive on every
constituent benchmark: \(+2.3\) on MMSrch-Plus, \(+11.1\) on HR-MMSrch,
\(+7.5\) on BC-VL, and \(+22.4\) on MMSrch. Thus, the online QA--Solver loop
continues to improve a substantially stronger Solver, rather than being
specific to small models.

Table~\ref{tab:qwen36-27b-scaling} additionally provides context from reported
27--32B multimodal deep-search agents. The external rows use earlier
Qwen3-VL backbones, including dense 32B and 30B-A3B MoE variants, whereas our
matched experiment uses Qwen3.6-27B. These values are not a controlled
ranking: benchmark subsets, tool stacks, search budgets, and judges differ
across papers. We therefore omit their aggregate scores and make the scaling
claim only from the local, matched base-versus-\methodname{} comparison.

\vspace{-2mm}
\section{Conclusion}
\label{sec:conclusion}
\vspace{-2mm}

We introduced \methodname{}, an online QA--Solver co-training framework for
improving multimodal deep-search agents without a pre-generated multi-hop
corpus or an additional task-specific SFT stage. A trainable QA Generator retrieves evidence and builds
multi-hop questions. The Solver's group rollout outcomes then update both
policies with separate GRPO objectives: the QA Generator learns to propose
questions that are difficult but solvable for the current Solver, while the
Solver learns from the resulting mixed-outcome groups. 
Across six multimodal deep-search evaluation benchmarks,
\methodname{} improves a Qwen3-VL-\(8\)B Solver from \(36.91\%\) to
\(51.58\%\) mean accuracy under the same agent workflow, while also improving
a newer Qwen3.6-\(27\)B backbone. 
The experiments show that a valid static QA policy becomes increasingly
mismatched as the Solver changes, whereas updating the QA policy restores a
larger supply of useful training groups. The reward ablation further shows
that questions should be challenging without being pushed too close to
universal failure. These results support verified online task generation as a
practical route to self-improving multimodal deep-search agents. Extending the
harness to richer tool environments while reducing multi-tool rollout cost is
a promising direction for future work.

\subsection*{AI Use Statement}
We used generative AI tools for literature organization, plotting and
analysis code, and experiment-orchestration and debugging scripts. The research
questions, method design, interpretation, and conclusions are the authors' own;
the authors reviewed and tested AI-assisted code and verified reported results
against experiment outputs and raw logs. Generative models also have substantive
roles in the method and evaluation: Qwen models instantiate the trainable QA
Generator and Solver, a frozen Claude Sonnet~$5$ endpoint supports model-assisted
harness checks, \texttt{gpt-5-mini} adjudicates answers not resolved by
deterministic matching, and Claude~$4.8$ Opus is used for the external QA-quality
audit. Proprietary models are additionally evaluated as baselines. These uses
are described in the corresponding method, experiment, and appendix sections.
We take responsibility for the final text, claims, code, and experimental
artifacts reported in this work.

\subsection*{Ethics Statement}
This work uses publicly released models and VQA datasets together with publicly
accessible web search results. The reported manual audit evaluates model
outputs rather than participant behavior or personal attributes. We do not
intentionally target private or sensitive information, although live web
retrieval can surface personal, copyrighted, biased, outdated, or otherwise
unreliable material. The fixed harness and answer adjudication reduce but do not
eliminate these risks. Improved deep-search agents also carry dual-use risks by
making large-scale information collection and synthesis easier. Deployments in
high-stakes settings should therefore use appropriate access controls, source
review, and human oversight. We report the main run's compute and external-call
footprint in Appendix~\ref{app:data-compute}.

\subsection*{Reproducibility Statement}
Sections~\ref{sec:method} and~\ref{sec:experimental-setup} specify the model,
data, tool-use protocol, rewards, and optimization setup. The appendix provides
the QA state and entity definitions, ordered harness checks, Solver interfaces,
token-level masking rules, scoring procedure, training configuration, and
verified workload and resource accounting. Training uses publicly released
Qwen backbones and seed contexts drawn from public VQA datasets. Exact replay of
live retrieval and proprietary model-assisted checks may vary with web content
and service versions; we therefore separate deterministic checks from
model-assisted checks and report their roles and failure behavior. These
disclosures support protocol-level reproduction while making the remaining
dependency on time-varying external services explicit.

\bibliography{iclr2027_conference}
\bibliographystyle{iclr2027_conference}

\clearpage
\appendix

\section{Detailed Agent Protocols}
\label{app}

\subsection{Bridge-Entity Definition and Generator State}
\label{app:entity-state}

Following the entity convention used to construct our seed pool
\citep{peng2026mta}, we define a \emph{bridge entity} at every non-final hop as
a single proper-name answer that denotes one unambiguous real-world referent,
is supported by the retrieved evidence, and can anchor the next retrieval step. This covers named
referents from the categories summarized in
Table~\ref{tab:bridge-entity-taxonomy}.

\begin{table*}[t]
\centering
\caption{Bridge-entity taxonomy used by the QA construction system, adapted
from the seed-selection convention of MTA-Agent~\citep{peng2026mta}. The
taxonomy constrains intermediate answers, which must support another retrieval
step; a final-hop answer may instead be a short non-entity fact.}
\label{tab:bridge-entity-taxonomy}
\footnotesize
\setlength{\tabcolsep}{5pt}
\begin{tabular}{@{}p{0.18\textwidth}p{0.48\textwidth}p{0.24\textwidth}@{}}
\toprule
Category & Included referents & Valid bridge-answer examples \\
\midrule
People & Uniquely named public, historical, scientific, or athletic figures
& Ada Lovelace; Serena Williams \\
Organizations & Named institutions, companies, universities, government
bodies, or sports teams & UNESCO; FC Barcelona \\
Locations and landmarks & Specific geographic, natural, or architectural
referents & Nairobi; Machu Picchu \\
Products & Commercially named devices, vehicles, software, or other products
& PlayStation~$5$; Toyota Prius \\
Events & Named sporting, historical, academic, or cultural events
& Cannes Film Festival; $2016$ Summer Olympics \\
Creative works & Titled films, books, artworks, television series, or music
& \textit{Spirited Away}; \textit{The Starry Night} \\
Science and technology & Named species, compounds, devices, models, or datasets
& Tyrannosaurus rex; BERT \\
Medical and health & Named diseases, conditions, or other specific medical
entities & Alzheimer's disease; dengue fever \\
Business and finance & Named brands, firms, indexes, or financial instruments
& Patagonia; Nasdaq-$100$ \\
Space and astronomy & Named celestial objects, missions, telescopes, or
astronomical events & Voyager~$1$; Andromeda Galaxy \\
\bottomrule
\end{tabular}
\end{table*}

A multi-entity answer, unsupported alias, descriptive phrase, generic
category, action, or bare quantity is not a valid bridge entity. The final hop
need not support another extension: when the final-hop quantity option is
enabled, it may instead end in an exact count, measurement, date, or other
short factual answer.

The policy proposes the complete answer string \(a^h\), but it does not create
the two entity representations used by the next turn. After accepting a
non-final hop, the fixed harness derives \(e^h\), a short canonical form used
inside the next subquestion, and \(\tilde e^h\), a fully disambiguated form used
inside the next search query. For example, an accepted answer ``Window Rock,
Arizona'' is represented as \(e^h=\) ``Window Rock'' and
\(\tilde e^h=\) ``Window Rock, Arizona, the capital of the Navajo Nation.''
This separation keeps questions readable while making retrieval queries
unambiguous.

At hop \(h\), the policy state contains the seed image, the running merged
question \(\bar q^{h}\), \(e^h\), \(\tilde e^h\), the accepted answer history,
the latest harness observation, and the remaining hop/turn budget. Its two
actions and the corresponding fixed-harness responses are summarized below.

\begin{table*}[t]
\centering
\caption{QA Generator state transitions. Harness observations are appended as
user turns and are always masked from the policy loss.}
\label{tab:qa-state-action}
\footnotesize
\setlength{\tabcolsep}{4pt}
\begin{tabular}{@{}p{0.16\textwidth}p{0.32\textwidth}p{0.43\textwidth}@{}}
\toprule
Stage & Trainable QA Generator output & Fixed harness observation \\
\midrule
Search & \texttt{action}, \texttt{thought}, \texttt{tool=search}, and
\texttt{query}; the query must contain \(\tilde e^h\). & Retrieved and filtered
evidence chunks, followed by a request to propose one supported next-hop QA
pair. \\
Proposal & \texttt{action}, \texttt{thought}, \texttt{question},
\texttt{answer}, \texttt{evidence}, and \texttt{source}; the question must
depend on \(e^h\). & Either an acceptance containing \(\bar q^{h+1}\),
\(e^{h+1}\), and \(\tilde e^{h+1}\), or a rejection diagnosis that preserves
the last accepted state. \\
\bottomrule
\end{tabular}
\end{table*}

\subsection{A Recorded QA Generation Trajectory}
\label{app:qa-example}

We report one completed \(5\)-hop trajectory from a recorded QA Generator
rollout. Its final question received \(3/8\) correct Solver rollouts, making it
challenging but not uniformly unsolved. Each displayed query, evidence excerpt,
accepted model proposal, and harness merge is copied from the trace. We show
only the accepted path; rejected candidates are omitted and do not receive QA
policy credit. Construction-time source URLs are also omitted because the
Solver does not receive them.

\subsubsection{Monument Valley \(\rightarrow\) Window Rock Navajo Tribal Park}

\noindent
\begin{minipage}[t]{0.24\linewidth}
\vspace{0pt}\centering
\includegraphics[width=\linewidth,height=2.0cm,keepaspectratio]
{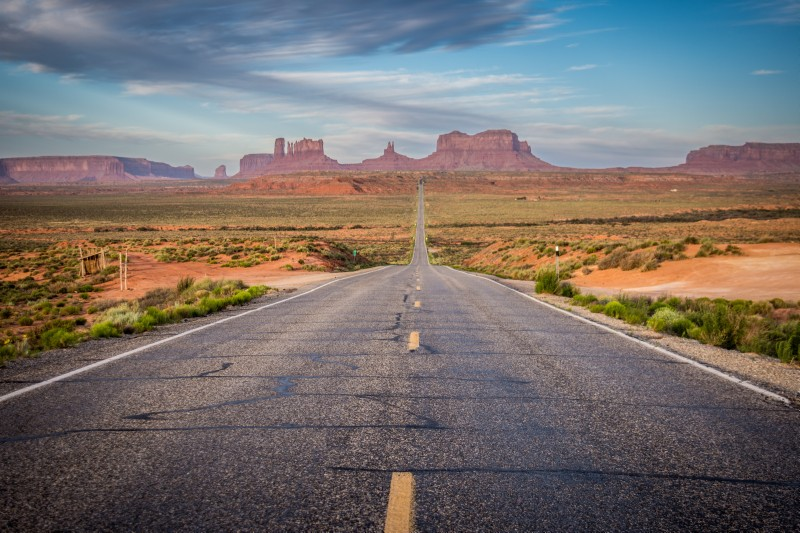}\\[-1mm]
\scriptsize\textit{Seed image}
\end{minipage}\hfill
\begin{minipage}[t]{0.72\linewidth}
\vspace{0pt}\textbf{Seed QA.} ``Which protected area is the place in the image
located in?'' \(\rightarrow\) ``Monument Valley Navajo Tribal Park.''\\[-1mm]
\textbf{Recorded Solver group.} \(3/8\) correct rollouts.
\end{minipage}\par\vspace{1.5mm}

{\small
\begin{enumerate}[labelindent=1em,labelwidth=2.25em,labelsep=.5em,leftmargin=3.75em,align=left,itemsep=5pt,topsep=3pt]
  \item[\(h=2\)] \textbf{Query.} ``Monument Valley Navajo Tribal Park,
  located in which U.S. state?''
  \textbf{Evidence \(E_1\).} ``The valley is considered sacred by the Navajo
  Nation, which administers the area as Monument Valley Navajo Tribal Park.''
  \textbf{Accepted QA.} ``Which Native American nation administers Monument
  Valley Navajo Tribal Park?'' \(\rightarrow\) ``Navajo Nation.''
  \textbf{Merge.} ``Which Native American nation administers the protected area
  where the place in the image is located?''

  \item[\(h=3\)] \textbf{Query.} ``Navajo Nation, the Native American nation
  that administers Monument Valley Navajo Tribal Park, what is its official
  capital or administrative center?''
  \textbf{Evidence \(E_1\).} ``The seat of government is located in Window
  Rock, Arizona.''
  \textbf{Accepted QA.} ``What is the official capital or administrative
  center of the Navajo Nation?'' \(\rightarrow\) ``Window Rock, Arizona.''
  \textbf{Merge.} ``What is the official capital or administrative center of
  the Native American nation that administers the protected area where the
  place in the image is located?''

  \item[\(h=4\)] \textbf{Query.} ``Window Rock, Arizona, the capital of the
  Navajo Nation, what is the name of its official tribal government building?''
  \textbf{Evidence \(E_4\).} ``This small park established sometime around
  \(1936\), features the graceful red sandstone arch for which the capital is
  named Ts\'egh\'ahoodz\'an\'i (Window Rock).''
  \textbf{Accepted QA.} ``What is the name of the red sandstone arch for which
  the capital of Window Rock, Arizona, is named?'' \(\rightarrow\)
  ``Ts\'egh\'ahoodz\'an\'i.''
  \textbf{Merge.} ``What is the name of the red sandstone arch for which the
  capital of the Native American nation that administers the protected area
  where the place in the image is located is named?''

  \item[\(h=5\)] \textbf{Query.} ``Ts\'egh\'ahoodz\'an\'i, the red sandstone
  arch in Window Rock, Arizona, capital of the Navajo Nation, what is the
  official name of the park where it is located?''
  \textbf{Evidence \(E_1\).} ``This small park established sometime around
  \(1936\), features the graceful red sandstone arch for which the capital is
  named Ts\'egh\'ahoodz\'an\'i (Window Rock). The park, officially known as
  Window Rock Navajo Tribal Park \& Veterans Memorial, is located adjacent to
  the Navajo Nation government offices in Window Rock, Arizona.''
  \textbf{Accepted QA.} ``What is the official name of the park where the red
  sandstone arch Ts\'egh\'ahoodz\'an\'i is located?'' \(\rightarrow\) ``Window
  Rock Navajo Tribal Park \& Veterans Memorial.''
  \textbf{Final merged task.} ``What is the official name of the park where the
  red sandstone arch, for which the capital of the Native American nation that
  administers the protected area where the place in the image is located is
  named, is itself located?'' \(\rightarrow\) ``Window Rock Navajo Tribal Park
  \& Veterans Memorial.''
\end{enumerate}
}

\newpage
\section{QA Harness Checks and Feedback}
\label{app:qa_harness}

The QA harness is fixed throughout co-training. Its validation and merging
stages follow MTA-Agent~\citep{peng2026mta}; our adaptation exposes structured
accept/reject feedback to the online QA policy and masks rejected action spans.
The harness uses retrieval together with fixed programmatic and model-based
checks, but its parameters are not optimized and no gradient flows through it.
The main run uses Serper \footnote{\href{https://serper.dev/}{https://serper.dev/}} for text retrieval and routes all model-assisted
harness calls through the same frozen Claude Sonnet~\(5\) endpoint on AWS
Bedrock. Calls use a \(60\)-second timeout and at most \(3\) API attempts. The
ordered checks are detailed in Table~\ref{tab:harness-checks}.

\begin{table*}[t]
\centering
\caption{Fixed QA-harness execution order. Model-call failure behavior is
reported explicitly; verifier availability is monitored during training.}
\label{tab:harness-checks}
\scriptsize
\setlength{\tabcolsep}{3.5pt}
\begin{tabular}{@{}cp{0.18\textwidth}p{0.42\textwidth}p{0.30\textwidth}@{}}
\toprule
\# & Stage & Operation & Failure behavior \\
\midrule
\(1\) & Retrieve and filter & Text search returns at most \(6\) results. The
filter may read up to \(6{,}000\) characters per result and exposes at most
\(6\) evidence chunks of \(1{,}200\) characters each to the QA policy; raw
text retained internally is capped at \(8{,}000\) characters per result. & An
empty/failed search rejects the attempt. If the filter call is unavailable or
malformed, bounded raw snippets are used; an explicit empty filtered set
rejects the attempt. \\
\\
\(2\) & Structural and rarity gates & Require nonempty question, answer, and
evidence; reject a repeated answer or an answer appearing in more than \(3\)
retrieved chunks. & Deterministic rejection. \\
\\
\(3\) & Entity continuity & Compare the proposed question against both the
short current entity \(e^h\) and its disambiguated form \(\tilde e^h\), checking
that the question truly depends on the same referent. & A valid negative verdict
rejects; an unavailable or malformed model verdict does not veto the proposal.
\\
\\
\(4\) & Proposal quality & Rule-based and model-assisted checks test the
proposal's supplied evidence for support, and test uniqueness, temporal
stability, entity necessity, stand-alone wording, and non-final answer
linkability. & Rule violations reject deterministically; an unavailable or
malformed model verdict does not veto the proposal. \\
\\
\(5\) & Merge and normalize & Replace the previous explicit entity in the new
question with the running description; return the merged question together
with \(e^{h+1}\) and \(\tilde e^{h+1}\). & Missing or unparsable merged output
rejects the attempt. \\
\\
\(6\) & Merged-QA coherence & Check whether the merged question remains
self-contained and has the proposed answer as its unique intended resolution.
& A valid negative verdict rejects; an unavailable or malformed verdict does
not veto the proposal. \\
\\
\(7\) & Parametric direct-answer gate & Ask the frozen helper to answer the
merged question without web access. The proposal is rejected when its expected
answer appears in that response under case-insensitive substring matching. & A
failed helper call does not veto the proposal. \\
\bottomrule
\end{tabular}
\end{table*}


For an accepted proposal, the harness appends a user-turn observation
containing \(\bar q^{h}\), \(a^{h}\), and an acceptance message.  This is
the state used for the next Generator decision.  A rejected proposal instead
produces diagnostic feedback, allowing another attempt while preserving the
previous accepted state.  Both types of feedback remain visible in the
context.  However, only the search action and QA proposal associated with an
accepted hop receive the QA policy-gradient signal. Tokens from rejected
attempts are masked from the QA loss. 

\subsection{Recorded Harness Feedback}
\label{app:qa-feedback-examples}

The following boxes reproduce the decision-bearing content returned to the QA
Generator in recorded rollouts, with field names normalized for clarity. We
omit generic action-schema and budget instructions, but retain the query,
evidence, validation decision, and updated state. Each message is appended as a
user turn: it is context for the next policy decision rather than a training
target.

\paragraph{Search feedback.}
The harness returns filtered
evidence and switches the policy to the next-hop proposal stage. The following
is from the recorded trajectory in Appendix~\ref{app:qa-example}.

\begin{tcolorbox}[colback=gray!4,colframe=gray!60,boxrule=.35pt,arc=1mm,
  left=1mm,right=1mm,top=.7mm,bottom=.7mm]
\footnotesize
\texttt{[TOOL\_FEEDBACK]} \texttt{search} succeeded for Hop \(3/5\).\\
\textbf{Query:} \emph{Navajo Nation, the Native American nation that
administers Monument Valley Navajo Tribal Park, what is its official capital or
administrative center?}\\
\textbf{Filtered evidence (excerpt):} \emph{The seat of government is located
in Window Rock, Arizona.}\\
\texttt{Now propose ONE supported sub-question using only this evidence.}
\end{tcolorbox}

\paragraph{Accepted-hop feedback.}
Once a proposal passes validation and merging, the harness returns the
accepted QA pair, the merged question, and the entity to explore next.

\begin{tcolorbox}[colback=green!4,colframe=green!45!black,boxrule=.35pt,arc=1mm,
  left=1mm,right=1mm,top=.7mm,bottom=.7mm]
\footnotesize
\texttt{[SUCCESS]} Hop \(3/5\) completed.\\
\textbf{Used result:} Evidence \(1\)\\
\textbf{Accepted QA proposal:} ``What is the official capital or
administrative center of the Navajo Nation?'' \(\rightarrow\) ``Window Rock,
Arizona.''\\
\textbf{Merged question (after this hop):} ``What is the official capital or
administrative center of the Native American nation that administers the
protected area where the place in the image is located?''\\
\textbf{Next entity to explore:} ``Window Rock, Arizona.''
\end{tcolorbox}

\paragraph{Rejected-hop feedback.}
If validation fails, the harness preserves the last accepted state, provides a
specific diagnosis, and requests a different search angle. The following
recorded White Tower proposal is rejected as too generic.

\begin{tcolorbox}[colback=red!3,colframe=red!55!black,boxrule=.35pt,arc=1mm,
  left=1mm,right=1mm,top=.7mm,bottom=.7mm]
\footnotesize
\texttt{[FAILED]} Hop \(3/5\) -- your sub-question proposal failed verification.\\
\textbf{Root cause:} \texttt{qa\_verification\_failed:}\\[-1pt]
\hspace*{1.5em}\texttt{vague\_or\_generic\_question}\\
\textbf{Recommendation:} Try a different search angle and propose a more
specific, evidence-supported question.\\
\textbf{Rejected proposal:} ``What architectural style does the White Tower at
the Tower of London, England, exemplify?'' \(\rightarrow\) ``Norman style
architecture.''\\
\textbf{Reason:} Asking ``what architectural style'' is a generic category
question, and the answer is not unique to the White Tower.\\
\textbf{Preserved state:} The current entity is still ``White Tower.'' Choose a
different query.
\end{tcolorbox}

The failed search--proposal span remains visible in context but is masked from
the QA loss. Only a search action and proposal that form an accepted hop
contribute to the QA GRPO update.

\subsection{External Quality Audit of Generated QA Tasks}
\label{app:qa-quality-audit}

The rollout-success target alone does not guarantee that a QA Generator avoids
ambiguous or otherwise shortcut tasks. We therefore conduct an external
LLM-based quality audit with Claude $4.8$ Opus, separately from the QA harness
and Solver. For each of $Q_0$ and $Q_{50}$, the audit assesses $512$ generated
tasks under the same fixed rubric. The evaluator inspects each final task and
its recorded construction trace, and returns a binary judgment for every
criterion below.

\begin{table}[t]
\centering
\caption{External QA-quality audit of generated tasks. Claude $4.8$ Opus
audits $512$ tasks from each QA checkpoint under the same fixed rubric. Values
are the percentage of tasks judged to satisfy each criterion; higher is better.}
\label{tab:qa-quality-audit}
\small
\setlength{\tabcolsep}{4pt}
\begin{tabular}{@{}lcc@{}}
\toprule
Audit criterion & $Q_0$ & $Q_{50}$ \\
\midrule
Valid generated task & $90.0\%$ & \textbf{$92.0\%$} \\
Evidence / chain support & $70.3\%$ & \textbf{$80.3\%$} \\
Answer unique and stable & $81.3\%$ & \textbf{$83.0\%$} \\
Self-contained question & $90.2\%$ & \textbf{$93.5\%$} \\
No answer or evidence leakage & $87.9\%$ & \textbf{$89.8\%$} \\
Requires the seed image & \textbf{$73.8\%$} & $73.0\%$ \\
\bottomrule
\end{tabular}
\end{table}

\paragraph{Audit criteria.}
\emph{Valid generated task} requires a usable final question--answer pair.
\emph{Evidence / chain support} requires that the cited evidence supports each
accepted hop and that the merged chain preserves the intended answer.
\emph{Answer unique and stable} rejects plausible alternative answers or
time-sensitive answers without an explicit date. \emph{Self-contained} checks
that the final question can be interpreted without the hidden construction
state. \emph{No answer or evidence leakage} checks that intermediate answers
and minting evidence do not appear in the final task. Finally, \emph{requires
the seed image} tests whether the image remains necessary to answer the final
question.

As Table~\ref{tab:qa-quality-audit} shows, all five structural quality checks
increase from $Q_0$ to $Q_{50}$: validity, evidence/chain support, answer
uniqueness, self-containedness, and anti-leakage. In particular,
evidence/chain support rises by \(10.0\) percentage points and
self-containedness by \(3.3\) points. Image dependence is nearly unchanged,
at $73.8\%$ for $Q_0$ and $73.0\%$ for $Q_{50}$. This audit does not rule out
every possible shortcut or establish that every mixed-outcome group tests the
intended reasoning capability. It does, however, find no detectable quality
degradation from QA training in the direct signatures of ambiguity,
unsupported chains, answer leakage, or image irrelevance.

\subsection{Learning Boundary of the Fixed Harness}
\label{app:harness-boundary}

The harness is a fixed task-construction environment, not a second policy that
generates the training questions.  It contains retrieval backends,
programmatic filters, and fixed model-assisted checks, all held unchanged
throughout a run.  In particular, it never selects the next search action or
its arguments, and it does not author the candidate single-hop QA pair: these
are outputs of the trainable QA Generator.  Table~\ref{tab:harness-boundary}
separates the responsibilities and learning signals.

\begin{table}[t]
\centering
\caption{Learning boundary in the online QA construction loop.}
\label{tab:harness-boundary}
\small
\setlength{\tabcolsep}{3pt}
\begin{tabular}{@{}p{0.24\linewidth}p{0.38\linewidth}p{0.27\linewidth}@{}}
\toprule
Component & Input $\rightarrow$ output & Learning signal \\
\midrule
QA Generator & State $\rightarrow$ tool and arguments, or proposed
$(q^h,a^h)$ with evidence citation & QA GRPO with scalar $r_i^{\mathrm{Q}}$ \\
Fixed harness & Tool request $\rightarrow$ bounded evidence; proposal
$\rightarrow$ accept/reject feedback and merged question & Fixed; no updated
parameters, reward, or gradient \\
Solver & Seed image and final task $\rightarrow$ $K$ tool-use
rollouts & Solver GRPO with rollout correctness $c_{ij}$ \\
\bottomrule
\end{tabular}
\end{table}

\paragraph{Token-level QA credit assignment.}
The harness may return filtered evidence, a diagnostic rejection, or a merged
question, but each is appended as an observation rather than a target
completion and is always masked from the QA loss.  For an accepted hop, the
entire Generator-produced assistant span---reasoning, tool selection, tool
arguments, and the proposed question, answer, and evidence citation---remains
trainable.  If the hop is rejected, the corresponding span is masked starting
from that attempt's tool selection; this masks its reasoning, arguments, and
proposal together, while keeping the content visible as context for the next
attempt.  A later rejection never retroactively masks earlier accepted hops.
If generation stops before the target depth but has accepted earlier hops, the
current merged partial QA task is retained and those accepted spans remain
trainable.  

\paragraph{Rewards and update order.}
For each generated task, the $K=8$ binary Solver outcomes are first reduced to
$p_i$.  With the main target $\tau=0.5$, the QA Generator receives the single
scalar Goldilocks reward
\[
 r_i^{\mathrm{Q}}=
 \begin{cases}
  1-2\lvert p_i-0.5\rvert, & 0<p_i<1,\\
  0, & p_i\in\{0,1\}.
 \end{cases}
\]
The $N_{\mathrm{QA}}=8$ candidate trajectories from one seed form one QA GRPO
group.  Consequently, a candidate with reward zero is not automatically
dropped: it can receive a negative relative advantage when other candidates
from that seed have higher reward.  The main $8$B run keeps only Solver groups
with $0<p_i<1$ and uses the individual binary correctness values $c_{ij}$ as
their terminal rewards; it enables neither a function-call format gate nor
hop-level partial credit.  Each iteration executes QA rollout, harness
construction, $K$ Solver rollouts per task, correctness and QA-reward
calculation, then the QA update and the Solver update.  Both updates therefore
use the same pre-update rollout batch; the QA update does not regenerate the
Solver data within that iteration.

\paragraph{Information isolation.}
The Solver receives the seed image and final merged question, but not the
Generator's intermediate QA pairs, their answers, minting evidence, or mint
source URLs. Answer-bearing mint URLs are recorded only as blocked retrieval
URLs. The image URL is used solely as a handle for image tools such as reverse
image search, OCR, and cropping; no source HTML URL or other seed provenance
metadata is provided to the Solver. Thus, model-assisted checks can control task quality, but they do
not provide the Solver's answer or share a trainable representation, optimizer,
or gradient with either policy.  The QA Generator sees only its harness
feedback and the scalar $r_i^{\mathrm{Q}}$, never Solver tokens, hidden states,
or gradients.  The two policies have separate actor parameters and optimizers;
the main run does not enable a KL/reference-policy loss for either policy.

\newpage
\section{Solver Interaction and Tool Interfaces}
\label{app:solver-interaction}

For each Solver input, a rollout starts from its seed image and final merged
question.  At every ReAct step, the Solver emits a
reasoning trace, an action, a
\texttt{should\_stop} decision, and a confidence score.  An action specifies a
tool, its arguments, a short action description, and the expected observation.
The observation server executes the tool call and appends the result to the
interaction history before the next decision.  The rollout finishes when the
Solver returns its final answer, when \texttt{should\_stop} is true with
confidence above \(0.7\), or when it reaches the \(T=10\)-step budget.  The
final answer is emitted in a designated boxed-answer field for evaluation.

For every question \(q_i\), we sample \(K=8\) complete rollouts.  We first
apply answer normalization and matching, and use a fixed semantic judge only
for unresolved aliases or paraphrases.  This yields the binary terminal
correctness \(c_{ij}\) of rollout \(j\).  Its group rate \(p_i\) determines
the Goldilocks score used by the QA Generator and whether the Solver group is
retained for GRPO.  Tool observations are part of the rollout context and are
masked from the policy loss.

\subsection{Tool Palette}
\label{app:tools}

The Solver uses the following \(6\) callable tools in the reported
multi-tool environment.

\begin{table*}[t]
\centering
\caption{Solver tool-call interfaces and representative responses appended to
the next Solver turn.  The first five response excerpts are from recorded
validation trajectories.  The cropping response is from a successful live tool
call and includes an image block in addition to its text metadata.}
\label{tab:appendix-tools}
\small
\setlength{\tabcolsep}{4pt}
\begin{tabular}{@{}p{0.18\textwidth}p{0.20\textwidth}p{0.54\textwidth}@{}}
\toprule
Tool & Arguments & Representative returned response \\
\midrule
Web search & \texttt{query} &
{\scriptsize\ttfamily Tool: web\_search\par Response: The building was
constructed to display and promote local products, commerce, and industry.} \\
\midrule
Web image search & \texttt{query} &
{\scriptsize\ttfamily Tool: image\_search\par Response: Search returned web
image results about Payette Lake near McCall, Idaho. A geographic source
identifies its outflow as the Payette River.} \\
\midrule
Web-page reader & \texttt{url} &
{\scriptsize\ttfamily Tool: content\_extraction\par URL:
en.wikipedia.org/wiki/Parc\_de\_Belleville\par Response: Page topic: Parc de
Belleville, an urban park in Paris.} \\
\midrule
Reverse image search & \texttt{image\_url}; optional \texttt{query} &
{\scriptsize\ttfamily Tool: reverse\_image\_search\par Response: The image
most strongly matches the Petit Trianon at the Ch\^ateau de Versailles. Visual
matches include photos and pages associated with this landmark.} \\
\midrule
OCR & \texttt{image\_url} &
{\scriptsize\ttfamily Tool: ocr\_tool\par Response: Text found in image:\par
Sign: KOMORA WARSZAWA XIX W. THE WARSZAWA CHAMBER (19TH CENTURY)} \\
\midrule
Image cropping & \texttt{image\_url}, \texttt{crop\_area} &
{\scriptsize\ttfamily Tool: image\_cropping\par Output: Reasoning: Crop
tightly around the central ruined dome-shaped building.\par Original size:
\((800,564)\); Crops \((1)\): ruined dome building | bbox=[\ldots]\par
[cropped image attached]} \\
\bottomrule
\end{tabular}
\end{table*}

\newpage
\section{Training and Evaluation Details}
\label{app:training-details}

\subsection{Seed-Pool Construction}
\label{app:seed-selection}

We select the one-hop starting contexts using the seed-filtering protocol of
MTA-Agent~\citep{peng2026mta}. A source item must have an image-dependent
question, a visually correct answer, and a single specific answer entity.
We discard the previously generated extensions, evidence, and
tool trajectory and retain only \((I,q^1,a^1)\).

The fixed pool contains
\(1{,}019\) contexts: \(398\) from InfoSeek, \(300\) from FVQA-train, \(190\)
from InfoVQA, and \(131\) from OK-VQA. Although the source records also store a
disambiguated answer field, the online data loader passes only \(I\), \(q^1\),
and \(a^1\) to the QA Generator; no source chain or answer metadata is used.
Consequently, the comparison to MTA-Agent starts from its filtered one-hop
contexts while requiring \methodname{} to construct every training-time
multi-hop task online.

\subsection{Main Co-training Configuration}

Both policies begin from separate copies of
\texttt{Qwen3-VL-8B-Instruct}.  The fixed seed pool combines the four public
one-hop VQA sources listed in Section~\ref{sec:evaluation-datasets}. Each seed
supplies an image and one question--answer pair. It contains neither multi-hop
questions nor retrieved evidence. The following configuration is used for the
main \(8\)B
co-training run.

\begin{table}[t]
\centering
\caption{Main \(8\)B co-training configuration.}
\label{tab:appendix-training-details}
\small
\setlength{\tabcolsep}{7pt}
\begin{tabular}{@{}lc@{}}
\toprule
Setting & Value \\
\midrule
Seed contexts per iteration & \(8\) \\
QA trajectories per seed & \(N_{\mathrm{QA}}=8\) \\
Target construction depth & \(5\) hops (\(1\) seed + \(4\) extensions) \\
Solver rollouts per QA input & \(K=8\) \\
Configured QA / Solver \texttt{max\_turns} & \(16\) / \(10\) \\
Search attempts / hop & \(3\) \\
Prompt / response token limits & \(8{,}192\) / \(24{,}576\) \\
Co-training iterations & \(50\) \\
Learning rate & \(10^{-6}\) \\
QA sampling temperature / top-\(p\) & \(1.0\) / \(0.95\) \\
Default QA target & \(\tau=0.5\) \\
\bottomrule
\end{tabular}
\end{table}

Each iteration therefore contains \(8\times 8=64\) QA-generation
trajectories and schedules \(8\times 8\times 8=512\) Solver rollout inputs.
The QA Generator and Solver use separate parameters, optimizers,
and GRPO updates; the main run does not enable a KL/reference-policy loss. The
only Solver-to-Generator signal is the scalar rollout-derived score; no
trajectory, hidden state, or gradient is shared between the two policies.

\subsection{Online Data and Workload Accounting}
\label{app:data-compute}

To make the online training budget explicit, we distinguish the fixed one-hop
seed pool from the QA tasks constructed during co-training. The main \(8\)B
run samples \(400\) seed-context draws from a fixed pool of \(1{,}019\) public
image-grounded VQA contexts: \(398\) InfoSeek, \(300\) FVQA-train, \(190\)
InfoVQA, and \(131\) OK-VQA examples. Each source item contains an image and
one-hop QA pair, with no retrieved evidence, multi-hop question, or tool
trajectory.

\begin{table}[t]
\centering
\small
\caption{Online data and workload accounting for the main \(8\)B run.}
\label{tab:appendix-data-compute}
\setlength{\tabcolsep}{3pt}
\begin{tabular}{@{}p{0.65\linewidth}r@{}}
\toprule
Quantity & Main \(8\)B run \\
\midrule
Seed-context draws & \(50\times8=400\) \\
QA-generation trajectories attempted & \(50\times8\times8=3{,}200\) \\
Scheduled Solver rollout inputs & \(3{,}200\times8=25{,}600\) \\
Pre-generated multi-hop QA corpus & \(0\) \\
Additional SFT updates & \(0\) \\
\bottomrule
\end{tabular}
\end{table}

\paragraph{Compute and service footprint.}
Table~\ref{tab:appendix-main-run-resources} reports the end-to-end resource
use of this same main $8$B run.

\begin{table*}[t]
\centering
\footnotesize
\caption{Verified allocation, timing, and external-service workload for the
main $8$B run.}
\label{tab:appendix-main-run-resources}
\setlength{\tabcolsep}{4pt}
\renewcommand{\arraystretch}{1.08}
\begin{tabular}{@{}p{0.18\textwidth}p{0.27\textwidth}p{0.19\textwidth}p{0.27\textwidth}@{}}
\toprule
\multicolumn{4}{c}{\textbf{Run allocation}} \\
\cmidrule{1-4}
Metric & Value & Metric & Value \\
\midrule
Completed iterations & $50$ & Hardware & $2$ nodes, $16$ NVIDIA H200 GPUs \\
End-to-end wall-clock & $16{:}21{:}48$ & Allocated compute & $32.727$ node-h; $261.813$ H200 GPU-h \\
Device memory & $140.401$ GiB/GPU & Peak allocated memory & QA: $128.757$ GiB; Solver: $130.512$ GiB \\
&& Peak reserved memory & QA: $133.992$ GiB; Solver: $133.996$ GiB \\
\midrule
\multicolumn{2}{c}{\textbf{Exclusive top-level timing}} &
\multicolumn{2}{c}{\textbf{External-service workload}} \\
\cmidrule(r){1-2}\cmidrule(l){3-4}
Rollout phase & $14.048$ h & QA search requests & $17{,}000$ ($16{,}299$ succeeded; $701$ failed) \\
Reward and group filtering & $0.017$ h & Recorded QA verifier calls & $41{,}411$ ($4$ fail-open) \\
QA / Solver optimization & $0.624$ h / $1.298$ h & Solver tool calls & $79{,}103$ \\
Initialization, checkpoints, and other overhead & $0.376$ h & Solver LLM-judge calls & $20{,}758$ ($20{,}599$ returned; $159$ failed) \\
\bottomrule
\end{tabular}
\end{table*}

\subsection{Answer Scoring and LLM Adjudication}
\label{app:llm-judge}

We evaluate the final Solver using the same ReAct interaction protocol as at
training time and report accuracy. For each benchmark item, we extract
the final boxed answer when present (otherwise the final response text) and
first apply a deterministic, normalized match against the set of acceptable
reference answers. This rule stage handles differences such as capitalization,
whitespace, and standard multiple-choice formatting.

Only when this match fails do we invoke the semantic judge,
\texttt{gpt-5-mini}. The judge receives the research question, all accepted
reference answers, and the model's extracted final answer. It is instructed to
decide whether the candidate conveys the same core information as a reference,
allowing harmless aliases, paraphrases, and formatting variation, and returns a
JSON verdict with a Boolean correctness field and a brief reason. A response is
counted as correct if either the rule stage or the semantic judge returns a
positive verdict. 

\paragraph{Human audit.}
To assess the reliability of LLM adjudication, we manually re-evaluated
\(100\) LLM-judged answers and found \(98\%\) agreement between the semantic
judge and human correctness labels.

\end{document}